\documentclass[11pt,a4paper]{article}
\usepackage[hyperref]{acl2020}
\usepackage{times}
\usepackage{latexsym}

\usepackage{url}
\usepackage{graphicx}
\usepackage{amsmath}
\usepackage{amssymb}
\usepackage{CJKutf8}
\usepackage{multirow}
\usepackage{float}
\usepackage{placeins}

\usepackage{microtype}

\aclfinalcopy 
\def\aclpaperid{221} 

\title{Enhancing Pre-trained Chinese Character Representation with Word-aligned Attention}

\author{Yanzeng Li, Bowen Yu, Mengge Xue, Tingwen Liu\thanks{~~Corresponding author}\\
	Institute of Information Engineering, Chinese Academy of Sciences \\
	School of Cyber Security, University of Chinese Academy of Sciences\\
	\texttt{\{liyanzeng, yubowen, xuemengge, liutingwen\}@iie.ac.cn}
}

\begin{document}
\maketitle
\begin{abstract}
Most Chinese pre-trained models take character as the basic unit and learn representation according to character's external contexts, ignoring the semantics expressed in the word, which is the smallest meaningful utterance in Chinese.
Hence, we propose a novel word-aligned attention to exploit explicit word information, which is complementary to various character-based Chinese pre-trained language models.
Specifically, we devise a pooling mechanism to align the character-level attention to the word level and propose to alleviate the potential issue of segmentation error propagation by multi-source information fusion.
As a result, word and character information are explicitly integrated at the fine-tuning procedure.
Experimental results on five Chinese NLP benchmark tasks demonstrate that our method achieves significant improvements against BERT, ERNIE and BERT-wwm.
\end{abstract}

\section{Introduction}\label{sec:intro}

Pre-trained language Models (PLM) such as ELMo \cite{ELMo}, BERT \cite{BERT}, ERNIE \cite{ERNIE}, BERT-wwm  \cite{BERTwwmreport} and XLNet \cite{XLNet} have been proven to capture rich language information from text and then benefit many NLP applications by simple fine-tuning, including sentiment classification~\cite{pang2002thumbs}, natural language inference~\cite{bowman-etal-2015-large}, named entity recognition~\cite{tjong2003introduction} and so on. 

Generally, most popular PLMs prefer to use attention mechanism \cite{vaswani2017attention} to represent the natural language, such as word-to-word self-attention for English.
Unlike English, in Chinese, words are not separated by explicit delimiters.
Since without word boundaries information, it is intuitive to model characters in Chinese tasks directly.
However, in most cases, the semantic of a single Chinese character is ambiguous.
\begin{CJK}{UTF8}{gbsn}
For example, the character ``拍'' in word ``球拍 (bat)'' and ``拍卖 (auction)'' has entirely different meanings.
\end{CJK}
Moreover, several recent works have demonstrated that considering the word segmentation information can lead to better language understanding, and accordingly benefits various Chinese tasks~\cite{wang-etal-2017-exploiting,li2018character, zhang2018chinese,gui2019cnn,gugugu}.

All these factors motivate us to expand the character-level attention mechanism in Chinese PLMs to represent the semantics of words \footnote{Considering the enormous cost of re-training a language model, we hope to incorporate word segmentation information to the fine-tuning process to enhance performance, and leave how to improve the pre-training procedure for a future work.}. 
To this end, there are two main challenges. 
(1) How to seamlessly integrate the segmentation information into character-based attention module of PLM is an important problem.
(2) Gold-standard segmentation is rarely available in the downstream tasks, and how to effectively reduce the cascading noise caused by Chinese word segmentation (CWS) tools \cite{li2019word} is another challenge. 

In this paper, we propose a new architecture, named
\textbf{M}ulti-source \textbf{W}ord \textbf{A}ligned Attention (MWA), to solve the above issues.
(1) Psycholinguistic experiments \cite{bai2008reading, meng2014landing} have shown that readers are likely to pay approximate attention to each character in one Chinese word. Drawing inspiration from such findings, we introduce a novel word-aligned attention, which could aggregate attention weight of characters in one word into a unified value with the mixed pooling strategy \cite{mixpooling}.
(2) For reducing segmentation error, we further extend our word-aligned attention with multi-source segmentation produced by various segmenters and deploy a fusion function to pull together their disparate outputs.
	As shown in Table~\ref{table:tokenizer}, different CWS tools may have different annotation granularity.
	Through comprehensive consideration of multi-granularity segmentation results, we can implicitly reduce the error caused by automatic annotation.

Extensive experiments are conducted on various Chinese NLP tasks including sentiment classification, named entity recognition, sentence pair matching, natural language inference and machine reading comprehension. 
The results and analysis show that the proposed method boosts BERT, ERNIE and BERT-wwm significantly on all the datasets \footnote{The source code of this paper can be obtained from \url{https://github.com/lsvih/MWA}.}.
\section{Methodology}\label{sec:method}

\begin{figure}
	\centering{\includegraphics[width = .32\textheight]{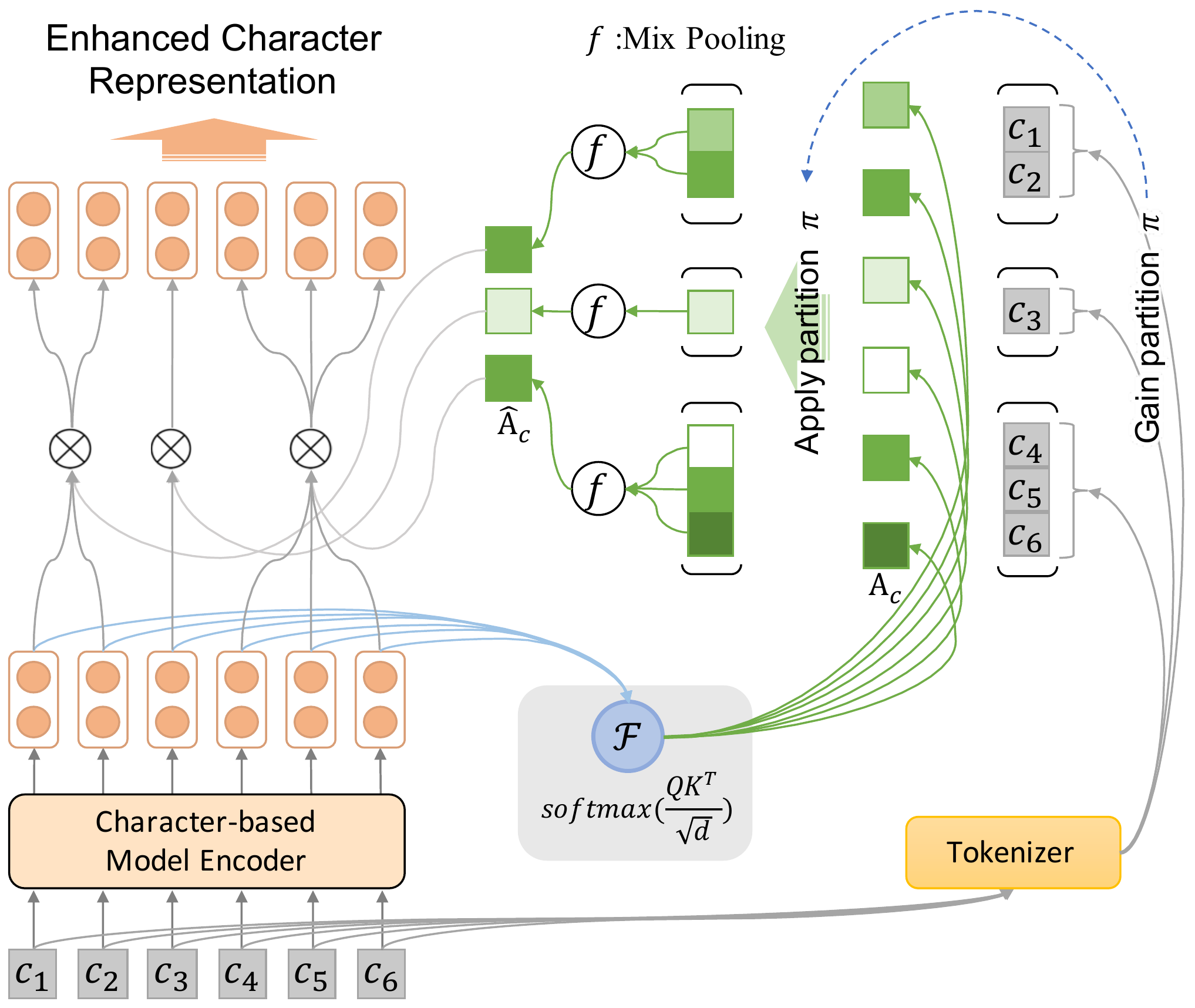}}
	\caption{Architecture of Word-aligned Attention}
	\label{fig:model}
	\vspace{-0.2cm}
\end{figure}

\subsection{Character-level Pre-trained Encoder}

The primary goal of this work is to inject the word segmentation knowledge into character-level Chinese PLMs and enhance original models.
Given the strong performance of deep Transformers trained on language modeling, we adopt BERT and its updated variants (ERNIE, BERT-wwm) as the basic encoder in this work, and the outputs from the last layer of encoder are treated as the character-level enriched contextual representations $\textbf{H}$.

\subsection{Word-aligned Attention} \label{sec:model}

Although character-level Chinese PLM has remarkable ability to capture language knowledge from text, it neglects the semantic information expressed in the word level. 
Therefore we apply a word-aligned layer on top of the encoder to integrate the word boundary information into the representation of characters with an attention aggregation module.

For an input sequence with $n$ characters $S=[c_1, c_2, ... , c_n]$, where $c_j$ denotes the $j$-th character, CWS tool $\pi$ is used to partition $S$ into non-overlapping word blocks:
\begin{equation}
\pi(S) = [w_1, w_2, ..., w_m], (m \leq n) 
\label{equ:word_sequence}
\end{equation}
where $w_i = \{c_s, c_{s+1}, ..., c_{s+l-1}\}$ is the $i$-th segmented word with a length of $l$ and $s$ is the index of $w_i$'s first character in $S$.
We apply self-attention operation with the representations of all input characters to get the character-level attention score matrix $\textbf{A}_c \in \mathbb{R}^{n \times n}$. It can be formulated as: 
\begin{gather}
\textbf{A}_c = \mathcal{F}(\textbf{H}) = \text{softmax}(\frac{(\textbf{K}\textbf{W}_k)(\textbf{Q}\textbf{W}_q)^T }{\sqrt{d}})
\end{gather}
where $\textbf{Q}$ and $\textbf{K}$ are both equal to the collective representation $\textbf{H}$ at the last layer of the Chinese PLM, $\textbf{W}_k \in \mathbb{R}^{d\times d}$ and $\textbf{W}_q \in \mathbb{R}^{d\times d}$ are trainable parameters for projection. 
While $\textbf{A}_c$ models the relationship between two arbitrarily characters without regard to the word boundary, we argue that incorporating word as atom in the attention can better represent the semantics, as the literal meaning of each individual character can be quite different from the implied meaning of the whole word, and the simple weighted sum in the character level may lose word and word sequence information. 

\begin{table}[t]
 \addtolength{\tabcolsep}{0.75pt}
 \renewcommand{\arraystretch}{1.3}
 \begin{CJK*}{UTF8}{gkai}
  \begin{tabular}{|c|r|c|c|c|c|c|c|c|c|} \hline
   \multicolumn{2}{|r|}{\textbf{Chinese}}&\multicolumn{8}{c|}{北京西山森林公园} \\
   \hline
   
   \parbox[t]{3mm}{\multirow{3}{*}{\rotatebox[origin=c]{90}{\textbf{Segmenter}}}}
   
   &thulac & \multicolumn{2}{c|}{北京} & \multicolumn{2}{c|}{西山} & \multicolumn{2}{c|}{森林} & \multicolumn{2}{c|}{公园}\\ 
   \cline{2-10}
   &ictclas & \multicolumn{2}{c|}{北京} & \multicolumn{1}{c|}{西} & \multicolumn{1}{c|}{山} & \multicolumn{2}{c|}{森林} & \multicolumn{2}{c|}{公园} \\ 
   \cline{2-10}
   &hanlp & \multicolumn{2}{c|}{北京} & \multicolumn{2}{c|}{西山} & \multicolumn{4}{c|}{森林公园} \\ \hline
  \end{tabular}\caption{Results of different popular CWS tools over ``北京西山森林公园 (Beijing west mount forest park)''.}
  \label{table:tokenizer}
 \end{CJK*}
\end{table}

To address this issue, we propose to align $\textbf{A}_c$ in the word level and integrate the inner-word attention.
For ease of exposition, we rewrite $\textbf{A}_c$ as $[\textbf{a}_c^1, \textbf{a}_c^2, ... ,\textbf{a}_c^n]$, where $\textbf{a}_c^i \in \mathbb{R}^n $ denotes the $i$-th row vector of $\textbf{A}_c$, that is, $\textbf{a}_c^i $ represents the attention score vector of the $i$-th character. 
Then we deploy $\pi$ to segment $\textbf{A}_c$ according to $\pi(S)$. For example, if $\pi(S) = [\{c_1, c_2\}, \{c_3\}, ...,\{c_{n-1}, c_{n}\}]$, then 
\begin{equation}
\pi(\textbf{A}_c) = [\{\textbf{a}_c^1, \textbf{a}_c^2\}, \{\textbf{a}_c^3\}, ...,\{\textbf{a}_c^{n-1}, \textbf{a}_c^{n}\}]
\end{equation} 

In this way, an attention vector sequence is divided into several subsequences and each subsequence represents the attention of one word. 
Then, motivated by the psycholinguistic finding that readers are likely to pay similar attention to each character in one Chinese word, we devise an appropriate aggregation module to fuse the inner-word attention. 
Concretely, we first transform $\{\textbf{a}_c^s,..., \textbf{a}_c^{s+l-1}\}$ into one attention vector $\textbf{a}_w^i$ for $w_i$ with the mixed pooling strategy \cite{mixpooling}~\footnote{Other pooling methods such as max pooling or mean pooling also works. 
Here we choose mixed pooling because it has the advantages of distilling the global and the most prominent features in one word at the same time.}.
Then we execute the piecewise upsampling operation over each $\textbf{a}_w^i$ to keep input and output dimensions unchanged for the sake of plug and play.
The detailed process can be summarized as:
\begin{align}
\label{equ:mix}
	\textbf{a}_w^i  & = \lambda\;\text{Maxpooling}(\{\textbf{a}_c^s,..., \textbf{a}_c^{s+l-1}\})  \\
						&\;\;\;\;+ (1 - \lambda)\;\text{Meanpooling} (\{\textbf{a}_c^s,..., \textbf{a}_c^{s+l-1}\}) \notag  \\
\label{equ:up}
	& \hat{\textbf{A}}_c[s:s+l-1] = \textbf{e}_l \otimes \textbf{a}_w^i
	\end{align}
where $\lambda \in R^1 $ is a weighting trainable variable to balance the mean and max pooling, $\textbf{e}_l=[1,...,1]^T$ represents a $l$-dimensional all-ones vector, $l$ is the length of $w_i$, $\textbf{e}_l \otimes \textbf{a}_w^i=[\textbf{a}_w^i,...,\textbf{a}_w^i]$ denotes the kronecker product operation between $\textbf{e}_l$ and $\textbf{a}_w^i$, $\hat{\textbf{A}}_c \in \mathbb{R}^{n \times n}$ is the aligned attention matrix.
Eqs. \ref{equ:mix} and \ref{equ:up} can help incorporate word segmentation information into character-level attention calculation process, and determine the attention vector of one character from the perspective of the whole word, which is beneficial for eliminating the attention bias caused by character ambiguity. 
Finally, we can obtain the enhanced character representation produced by word-aligned attention as follows: 
\begin{align}
\hat{\textbf{H}} = \hat{\textbf{A}}_c \textbf{V} \textbf{W}_v	
\end{align}
where $\textbf{V} = \textbf{H}$, $\textbf{W}_v \in \mathbb{R}^{d\times d}$ is a trainable projection matrix. Besides, we also use multi-head attention \cite{vaswani2017attention} to capture information from different representation subspaces jointly, thus we have $K$ different aligned attention matrices $\hat{\textbf{A}}_c^k (1\leq k\leq K)$ and corresponding representation $\hat{\textbf{H}}^k$.
With multi-head attention architecture, the output can be expressed as follows:
\begin{equation}
\label{equ:mh}
\overline{\textbf{H}} = \text{Concat}(\hat{\textbf{H}}^1,\hat{\textbf{H}}^2,...,\hat{\textbf{H}}^K)\textbf{W}_o
\end{equation}

\subsection{Multi-source Word-aligned Attention}

As mentioned in Section \ref{sec:intro}, our proposed word-aligned attention relies on the segmentation results of CWS tool $\pi$.
Unfortunately, a segmenter is usually unreliable due to the risk of ambiguous and non-formal input, especially on out-of-domain data, which may lead to error propagation and an unsatisfactory model performance.
In practice, the ambiguous distinction between morphemes and compound words leads to the cognitive divergence of words concepts, thus different $\pi$ may provide diverse $\pi(S)$ with various granularities.
To reduce the impact of segmentation error and effectively mine the common knowledge of different segmenters, it’s natural to enhance the word-aligned attention layer with multi-source segmentation inputs.
Formally, assume that there are $M$ popular CWS tools employed, we can obtain $M$ different representations $\overline{\textbf{H}}^1, ..., \overline{\textbf{H}}^M $ by Eq. \ref{equ:mh}. 
Then we propose to fuse these semantically different representations as follows:
\begin{equation}
\tilde{\textbf{H}} = \sum_{m=1}^{M} \tanh( \overline{{\textbf{H}}}^m\textbf{W}_g)
\end{equation}
where $\textbf{W}_g$ is a parameter matrix and $\tilde{\textbf{H}}$ denotes the final output of the MWA attention layer.

\section{Experiments}\label{sec:exp}

\subsection{Experiments Setup}

\begin{table*}\small
\centering
	\addtolength{\tabcolsep}{0px}
    \begin{tabular}{llccccccc}
    \hline
    \multicolumn{1}{c}{\multirow{2}*{Dataset}} & \multicolumn{1}{c}{\multirow{2}*{Task}} & \multicolumn{1}{c}{\multirow{2}*{Max length}} & \multicolumn{1}{c}{\multirow{2}*{Batch size}} & \multicolumn{1}{c}{\multirow{2}*{Epoch}} & \multicolumn{1}{c}{\multirow{2}*{lr$^*$}} & \multicolumn{3}{c}{Dataset Size} \\
	\cline{7-9}&&&&&& Train & Dev   & Test \\
    \hline
    ChnSentiCorp & \multirow{2}*{SC} & 256 & 16 & 3 & $3 \times 10^{-5}$ & 9.2K & 1.2K & 1.2K \\
    \cline{0-0}\cline{3-9} 
    weibo-100k &    & 128   & 64    & 2     & $2 \times 10^{-5}$ & 100K  & $\sim$10K  & 10K \\
    \hline
    ontonotes & NER & 256   & 16    & 5     & $3 \times 10^{-5}$ & 15.7K & 4.3K  & 4.3K \\
    \hline
    LCQMC & SPM  & 128   & 64    & 3     &  $3 \times 10^{-5}$ & $\sim$239K & 8.8K  & 12.5K \\
    \hline
    XNLI  & NLI   & 128   & 64    & 2     & $3 \times 10^{-5}$& $\sim$392K & 2.5K  & 2.5K \\
    \hline
    DRCD  & MRC  & 512   & 16    & 2     & $3 \times 10^{-5}$& 27K & 3.5K & 3.5K \\
    \hline
    \end{tabular}
    \caption{Summary of datasets and the corresponding hyper-parameters setting. Reported learning rates$^*$ are the initial values of BertAdam.}
    \label{tab:setting}
    \vspace{-0.2cm}
\end{table*}

To test the applicability of the proposed MWA attention, we choose three publicly available Chinese pre-trained models as the basic encoder: BERT, ERNIE, and BERT-wwm. 
In order to make a fair comparison, we keep \textbf{the same hyper-parameters} (such maximum length, warm-up steps, initial learning rate, etc.) as suggested in BERT-wwm \cite{BERTwwmreport}  for both baselines and our method on each dataset. 
\textbf{We run the same experiment for five times and report the average score} to ensure the reliability of results.
Besides, three popular CWS tools: thulac \cite{sun2016thulac}, ictclas \cite{zhang2003hhmm} and hanlp \cite{hanlp} are employed to segment sequence.

The experiments are carried out on five Chinese NLP tasks and six public benchmark datasets:

\textbf{Sentiment Classification (SC)}:  We adopt ChnSentiCorp\footnote{\url{https://github.com/pengming617/bert_classification}} and weibo-100k sentiment dataset\footnote{\url{https://github.com/SophonPlus/ChineseNlpCorpus/}} in this task. ChnSentiCorp dataset has about 10k sentences, which express positive or negative emotion. weibo-100k dataset contains 1.2M microblog texts and each microblog is tagged as positive or negative emotion.
 
 \textbf{Named Entity Recognition (NER)}: this task is to test model's capacity of sequence tagging. We use a common public dataset Ontonotes 4.0 \cite{weischedel2011ontonotes} in this task.
  
 \textbf{Sentence Pair Matching (SPM)}: We use the most widely used dataset LCQMC \cite{liu2018lcqmc} in this task, which aims to identify whether two questions are in a same intention.
 
 \textbf{Natural Language Inference (NLI)}: this task is to exploit the contexts of text and concern inference relationships between sentences. XNLI \cite{conneau2018xnli} is a cross-language language understanding dataset; we only use the Chinese language part of XNLI to evaluate the language understanding ability. And we processed this dataset in the same way as ERNIE \cite{ERNIE} did.

\textbf{Machine Reading Comprehension (MRC)}: MRC is a representative document-level modeling task which requires to answer the questions based on the given passages. DRCD \cite{DRCD} is a public span-extraction Chinese MRC dataset, whose answers are spans in the document.

We implement our model with PyTorch \cite{pytorch}, and all baselines are converted weights into PyTorch version.
All experiments employ modified Adam \cite{BERT} as optimizer with 0.01 weight decay and 0.1 warm-up ratio. All pre-trained models are configured to 12 layers and 768 hidden dimension. The detail settings are shown in Table \ref{tab:setting}.

\subsection{Experiment Results}

\begin{table*}
\small
	\centering
	\addtolength{\tabcolsep}{-1px}
	{
		\begin{tabular}{rllllll|l} \hline
			Task &    \multicolumn{2}{c}{SC}      &\multicolumn{1}{l}{NER}        &\multicolumn{1}{l}{SPM}    & \multicolumn{1}{l}{NLI}  & \multicolumn{2}{c}{MRC}\\
			\hline
			Dataset& ChnSenti$^{2,3}$ & weibo-100k$^2$ & Ontonotes$^4$ & LCQMC$^{2,3,4}$  & XNLI$^{1,2,3,4}$ & \multicolumn{2}{c}{DRCD$^{2,3}$ [EM$|$F1]} \\ \hline
			Prev. SOTA$^\dagger$& 93.1\shortcite{Glyce} &- &74.89\shortcite{LGN}&85.68\shortcite{MSEM}&67.5\shortcite{BiPMP}& 75.12\shortcite{lee2019crosslingual}&87.26\shortcite{lee2019crosslingual}\\\hline
			BERT & 94.72 & 97.31 & 79.18 & 86.50 & 78.19 & 85.57 & 91.16 \\
			+MWA & 95.34(+0.62) &98.14(+0.83)& 79.86(+0.68)&86.92(+0.42)&78.42(+0.23) &86.86(+1.29)&92.22(+1.06)\\ \hline
			BERT-wwm&94.38 & 97.36& 79.28 & 86.11 & 77.92 & 84.11& 90.46\\
			+MWA & 95.01(+0.63) & 98.13(+0.77)&\textbf{80.32}(+1.04)&86.28(+0.17)&78.68(+0.76) &87.00(+2.89)&92.21(+1.75)\\ \hline
			ERNIE& 95.17 & 97.30 & 77.74 & 87.27 & 78.04 &87.85& 92.85\\
			+MWA& \textbf{95.52}(+0.35) & \textbf{98.18}(+0.88)&78.78(+1.04)&\textbf{88.73}(+1.46)&\textbf{78.71}(+0.67) &\textbf{88.61}(+0.76)&\textbf{93.72}(+0.87)\\ \hline
		\end{tabular}
	}
	\caption{ Evaluation results regarding each model on different datasets. Bold marks highest number among all models.
Numbers in brackets indicate the absolute increase over baseline models. Superscript number$^{1,2,3,4}$ respectively represents that the corresponding dataset is also used by BERT \cite{BERT}, BERT-wwm \cite{BERTwwm, BERTwwmreport}, ERNIE \cite{ERNIE} and Glyce \cite{Glyce}, respectively.
The results of all baselines are produced by our implementation or retrieved from original papers, and we report the higher one among them.
The improvements over baselines are statistically significant ($p < 0.05$).
$^\dagger$ denotes the results of previous state-of-the-art models on these datasets without using BERT.
}
	\label{tab:result}
	\vspace{-0.2cm}
\end{table*}

Table \ref{tab:result} shows the performances on five classical Chinese NLP tasks with six public datasets.
Generally, our method consistently outperforms all baselines on all five tasks, which demonstrates the effectiveness and universality of the proposed approach.
Moreover, the Wilcoxon’s test shows that a significant difference ($p< 0.05$) exits between our model and baseline models.

In detail, on the two datasets of SC task, we observe an average of 0.53\% and 0.83\% absolute improvement in F1 score, respectively.
SPM and NLI tasks can also gain benefits from our enhanced representation.
For the NER task, our method obtains 0.92\% improvement averagely over all baselines. 
Besides, introducing word segmentation information into the encoding of character sequences improves the MRC performance on average by 1.22 points and 1.65 points in F1 and Exact Match (EM) score respectively.
We attribute such significant gain in NER and MRC to the particularity of these two tasks.
Intuitively, Chinese NER is correlated with word segmentation, and named entity boundaries are also word boundaries. 
Thus the potential boundary information presented by the additional segmentation input can provide better guidance to label each character, which is consistent with the conclusion in \cite{zhang2018chinese}.
Similarly, the span-extraction MRC task is to extract answer spans from document \cite{DRCD}, which also faces the same word boundary problem as NER, and the long sequence in MRC exacerbates the problem. 
Therefore, our method gets a relatively greater improvement on the DRCD dataset.

\subsection{Ablation Study}

\begin{table}[htp]
\small
 \centering 
 \begin{tabular}{lccc}
  \hline
  Model & BERT & BERT-wwm & ERNIE \\
  \hline
  Original& 92.06& 91.68& 92.61\\
  \hline
  +$1T$&92.37&92.22&93.42\\
  +$\text{WA}_{random}$&91.83&90.33&92.12\\
  \hline
  +$\text{WA}_{thulac}$ & 92.84 &92.73&93.89\\
  +$\text{WA}_{ictclas}$&93.05&92.90&93.75\\
  +$\text{WA}_{hanlp}$&92.91&93.21&93.91\\
  \hline
  +MWA&\textbf{93.59}&\textbf{93.72}&\textbf{94.21}\\
  \hline
 \end{tabular}
 \caption{ F1 results of ablation experiments on the DRCD dev set.}
 \label{tab:singlesa}
 \vspace{-0.2cm}
\end{table}

To demonstrate the effectiveness of our multi-source fusion method, we carry out experiments on the DRCD dev set with different segmentation inputs.
Besides, we also design two strong baselines by introducing a Transformer layer ($1T$) and a random tokenizer model ($\text{WA}_{random}$) to exclude the benefits from additional parameters.
As shown in Table \ref{tab:singlesa}, adding additional parameters by introducing an extra transformer layer can benefit the PLMs.
Compared with $1T$ and $\text{WA}_{random}$, our proposed word-aligned attention gives quite stable improvements no matter what CWS tool we use, which again confirms the effectiveness and rationality of incorporating word segmentation information into character-level PLMs.
Another observation is that employing multiple segmenters and fusing them together could introduce richer segmentation information and further improve the performance.
 
\subsection{Parameter Scale Analysis}

For fair comparison and demonstrating the improvement of our model is not only rely on more trainable parameters, we also conduct experiments on the DRCD dev set to explore whether the performance keeps going-up with more parameters by introducing additional transformer blocks on top of the representations of PLMs.

\begin{table}[htp]
\small
 \centering \addtolength{\tabcolsep}{-1px}
 \begin{tabular}{lcl}
  \hline
  Model & F1 & Param. Number \\
  \hline
  BERT-wwm& 91.68&110M\\ 
  BERT-wwm+$1T$&92.23&110M+7.1M\\
  BERT-wwm+$2T$&91.99&110M+14.2M\\
  BERT-wwm+$3T$&91.68&110M+21.3M\\
  \textbf{BERT-wwm+MWA}&93.72&110M+7.6M\\
  Robust-BERT-wwm-ext-large&94.40&340M\\
  \hline
 \end{tabular}
 \caption{Comparison on the DRCD dev set. The $n T$ denotes the number of additional transformer layers.}
 \label{tab:paramsize}
 \vspace{-0.2cm}
\end{table}

In Table \ref{tab:paramsize}, $+1T$ denotes that we introduce another one Transformer layer on top of BERT-wwm and $+2T$ means additional 2 layers, $M$ denotes million. 
As the experimental results showed, when the number of additional layers exceeds 1, the performance starts to decline, which demonstrates that using an extensive model on top of the PLM representations may not bring additional benefits. We can conclude that MWA doesn’t introduce too many parameters, and MWA achieves better performance than +1T under the similar parameter numbers.
Besides, we also make comparison with the current best Chinese PLM: Robust-BERT-wwm-ext-large \cite{BERTwwmreport}, a 24-layers Chinese PLM with 13.5 times more pre-training data and 3.1 times more parameters than BERT-wwm, experimental results show that our model can achieve comparable performance, which again confirms the effectiveness of incorporating word segmentation information into character-level PLMs.

\section{Conclusion}\label{sec:conclusion}

In this paper, we develop a novel Multi-source Word Aligned Attention model (referred as MWA), which integrates word segmentation information into character-level self-attention mechanism to enhance the fine-tuning performance of Chinese PLMs.
We conduct extensive experiments on five NLP tasks with six public datasets. 
The proposed approach yields substantial improvements compared to BERT, BERT-wwm and ERNIE, demonstrating its effectiveness and universality.
Furthermore, the word-aligned attention can also be applied to English PLMs to bridge the semantic gap between the whole word and the segmented WordPiece tokens, which we leave for future work.

\section*{Acknowledgement}

We would like to thank reviewers for their insightful comments. This work is supported by the Strategic Priority Research Program of Chinese Academy of Sciences, Grant No. XDC02040400.

\bibliographystyle{acl_natbib}
\bibliography{acl2020}

\end{document}